\documentclass{sig-alternate}
\usepackage{times}
\usepackage{epsfig}
\usepackage{graphics,graphicx, subfigure, color}
\usepackage{amsmath}
\usepackage{amssymb}
\newcommand{\figwidth}{0.8in}
\newcommand{\figheight}{0.8in}

\begin{document}

\title{Profile Based Sub-Image Search in Image Databases }

\numberofauthors{2}
\author{
\alignauthor
Vishwakarma Singh\\
       \affaddr{Department of Computer Science\\ University of California, Santa Barbara}\\
       \email{vsingh@ucsb.edu}
\alignauthor
Ambuj K. Singh\\
       \affaddr{Department of Computer Science\\ University of California, Santa Barbara}\\
      \email{ambuj@ucsb.edu}
}
\maketitle
\begin{abstract}
Sub-image search with high accuracy in natural images still remains a challenging problem. This paper proposes a new feature vector called \textit{profile} for a keypoint in a bag of visual words model of an image. The profile of a keypoint captures the spatial geometry of all the other keypoints in an image with respect to itself, and is very effective in discriminating true matches from false matches. Sub-image search using profiles is a single-phase process requiring no geometric validation, yields high precision on natural images, and works well on small visual codebook. The proposed search technique differs from traditional methods that first generate a set of candidates disregarding spatial information and then verify them geometrically. Conventional methods also use large codebooks. We achieve a precision of 81\%  on a combined data set of synthetic and real natural images using a codebook size of 500 for top-10 queries; that is 31\% higher than the conventional candidate generation approach.
\end{abstract}
\section{Introduction and Motivation}
Community contributed image sites (e.g. Flickr.com) and stock photo sites (e.g. Gettyimages.com) are witnessing an unprecedented growth in the recent decade. Search of images by example is one of the most common tasks performed on these data sets. A related task is sub-image retrieval~\cite{Philbin_2007,multi_scale,LUO_HIER}. It extends traditional full-image search by allowing users to select a region in an image and then, search for similar regions in a repository. Sub-image search is a very important tool for harnessing the potential of ever growing image repositories. Sub-image search as a tool is also being incorporated in online shopping (e.g. like.com) to help customers search products using images. It is also an important tool for analysis and study of biological and medical image datasets.

Community contributed repositories contain images of scenes or objects taken under varying imaging conditions. These images have affine, viewpoint, and photometric differences and also varying degrees of occlusion. Local descriptors like SIFT~\cite{LOWE_SIFT,comp_affine,MikolajczykPerfEval,Mikolajczyk02anaffine} are used in literature~\cite{Philbin_2007, wang_2008, Sukthanka_2004,sivic_videogoogle} with fair success to measure similarity between natural images. Images are scanned to detect keypoints~\cite{Harris}, covariant regions are extracted around each point, and finally a high dimensional local feature vector~\cite{LOWE_SIFT} representation of each region is obtained~\cite{comp_affine,Mikolajczyk02anaffine}. We know the spatial location, geometry of the covariant region, and high dimensional feature vector representation of each keypoint detected in an image. Thus, each image is transformed into a bag of feature vectors.
%
%
\begin{figure}[t]
  \centering
  \subfigure[Query]{\label{fig:harold}\includegraphics[height=0.6in,width=0.2\textwidth]{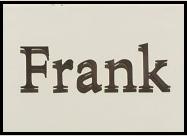}}
  \subfigure[Result]{\label{fig:harolds}\includegraphics[height=0.6in,width=0.2\textwidth]{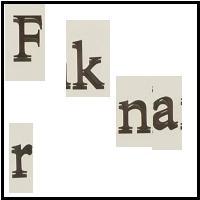}}
  \vspace*{-1mm}
  \caption{A pathological case for sub-image search using bag of words model and $L_p$ norm.  False match (b) is obtained in the top-5 results for query image (a) because spatial relationship is not considered.}
  \label{fig:path_1}
  \vspace*{-5mm}
\end{figure}

Researchers have pursued two-phase techniques to retrieve similar images using local descriptors. The first phase consists of candidate generation disregarding spatial relationships between keypoints and the second phase consists of geometric verification. Candidates are generated using two common approaches. One approach~\cite{Philbin_2007,sivic_videogoogle} transforms each image into an orderless bag of visual words. Feature vectors of all the images are clustered and each feature vector is assigned the symbol of its cluster called visual word. Thus, all the feature vectors of an image are transformed into visual words and an image is finally represented as a histogram of these visual words. This enables leveraging and extending existing text-retrieval techniques to image search. Distance between image histograms is computed using $L_p$ norm. A given query image is also transformed into a histogram of visual words and its top-k candidates are retrieved. Another approach~\cite{Sukthanka_2004,LOWE_SIFT} finds top-k nearest neighbors of each keypoint feature vector of the query image in the dataset of local feature vectors of all the images. It ranks images based on the total number of nearest neighbors detected in them and retains top-k images as candidates. Both approaches are made efficient by using indexing techniques~\cite{Arya,LSH}.

\begin{figure}
  \centering
  \subfigure[Query]{\label{fig:bird}\includegraphics[height=0.7in,width=0.2\textwidth]{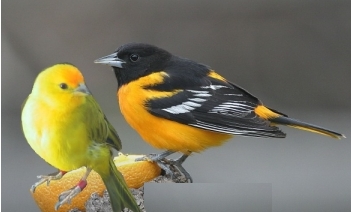}}
  \subfigure[Result]{\label{fig:birds}\includegraphics[height=1in,width=0.3\textwidth]{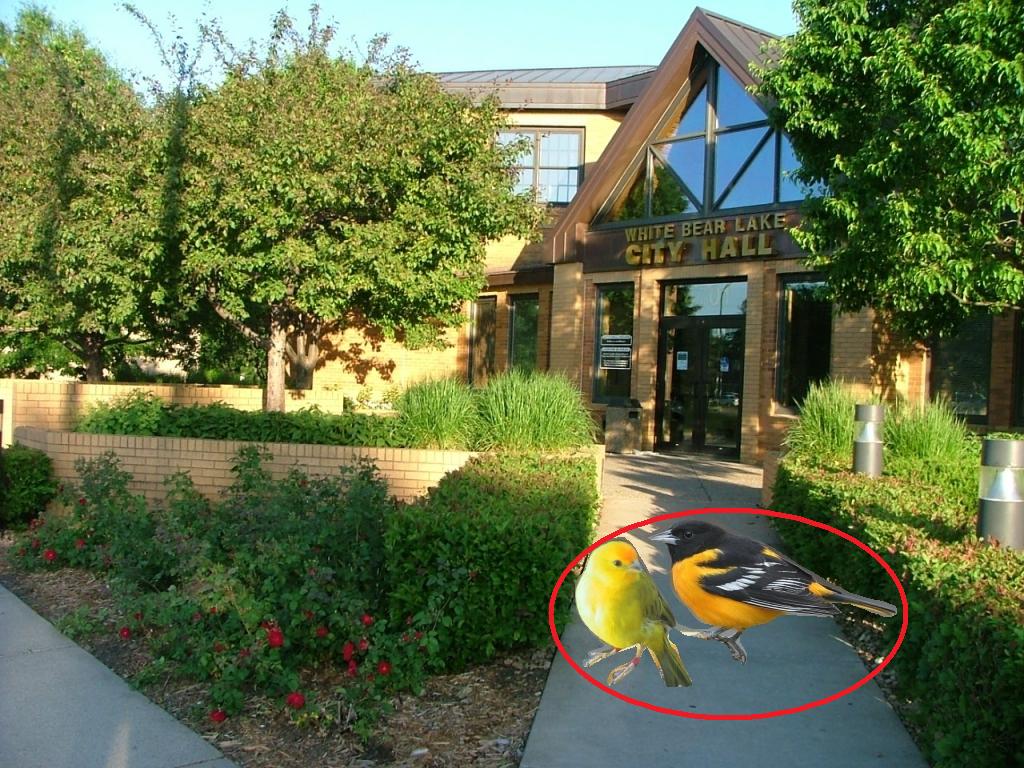}}
  \vspace*{0mm}
  \caption{A pathological case for sub-image search using bag of words model and $L_p$ norm. True match (b) is not retrieved because sub-image query (a) is a very small fraction of the actual image (b). }
  \label{fig:path_2}
  \vspace*{-2mm}
\end{figure}

We present two pathological results obtained by sub-image search using the first-phase methods as described previously. Figure~\ref{fig:harold} is a query sub-image and Figure~\ref{fig:harolds} is an image in the top-5 results retrieved from the dataset. We can see that the components of the query pattern are scattered randomly in Figure~\ref{fig:harolds} and it is not a meaningful match. A local descriptor like SIFT is computed just using neighborhood pixels around a keypoint. Similar keypoints will be computed in both the query and the given result image because result image contains all the components of the query. The only way to distinguish between these two images is to consider the spatial layout of the keypoints. These methods generate this false match because they do not take spatial relationships between keypoints into account. Figure~\ref{fig:bird} is another query sub-image and Figure~\ref{fig:birds} is a true match not returned in the top-5 results. An explanation of this anomaly is that the query region is a very small fraction of the whole image and therefore, random outliers score better than the true match when similarity is computed just using orderless bag of words representation of images. Both these examples motivate the necessity of using spatial relationships between matched points in a sub-image search to get high accuracy. Existing literature uses geometrical verification (Hough~\cite{LOWE_SIFT}, Ransack~\cite{Philbin_2007, Sukthanka_2004,Quack}) on the candidate images to find the best match or re-rank the top-$k$ candidates. Generating a small set of high quality candidates is very important for all the  conventional approaches to reduce the high cost of geometric verification. Geometric verification algorithms are iterative in nature and costly.

Sub-image search with high accuracy is very challenging and still remains an open problem. There is need of further development of better techniques to retrieve high quality images. Success of sub-image search tools in optimally utilizing image repositories for various applications is directly constrained by the accuracy of their search results. \textit{The focus of our work is to develop a feature vector and a sub-image similarity measure using this feature vector that yields high accuracy for sub-image search}.

In this paper, we present a new feature vector called \textit{profile}. We construct a concentric profile for each keypoint in a bag of visual words representation of an image (Section~\ref{sec:prof_creation}). A keypoint's profile approximately captures the spatial relationships of other keypoints in the image with respect to itself. Sub-image retrieval for a given query is a single-phase search of the best matching profile in the database (Section~\ref{sec:algorithm}). Our feature produces high quality results for sub-image search without geometric verification using a single-step search and a small visual codebook. We perform experiments to empirically validate our method in Section~\ref{sec:experiment}. We show that our profile achieves comparatively higher accuracy on the landmark dataset provided by Philbin et al.~\cite{Philbin_2007}. We also prepared a dataset which is a collection of natural and synthetic images. We used a small codebook of size $500$ for bag of words representation of images compared to thousands or even a million~\cite{Philbin_2007} used in the literature. A conventional user mostly views top-k results of a query~\cite{Browsing_Habit} returned by a search engine. Therefore, high accuracy in top-k results is really important. We specifically focus on empirically analyzing accuracy of top-k matching images retrieved using our feature vector. Our approach achieves 81\% precision on our dataset for top-$10$ results using code book size of $500$; this is 31\% higher than the precision of conventional methods.

The main contribution of our paper is the development of a robust feature vector for each keypoint that captures the spatial relationship of keypoints in an image, and a measure of similarity between the feature vectors for retrieving similar images.
\begin{figure*}[t]
  \centering
      \includegraphics[width=0.6\textwidth]{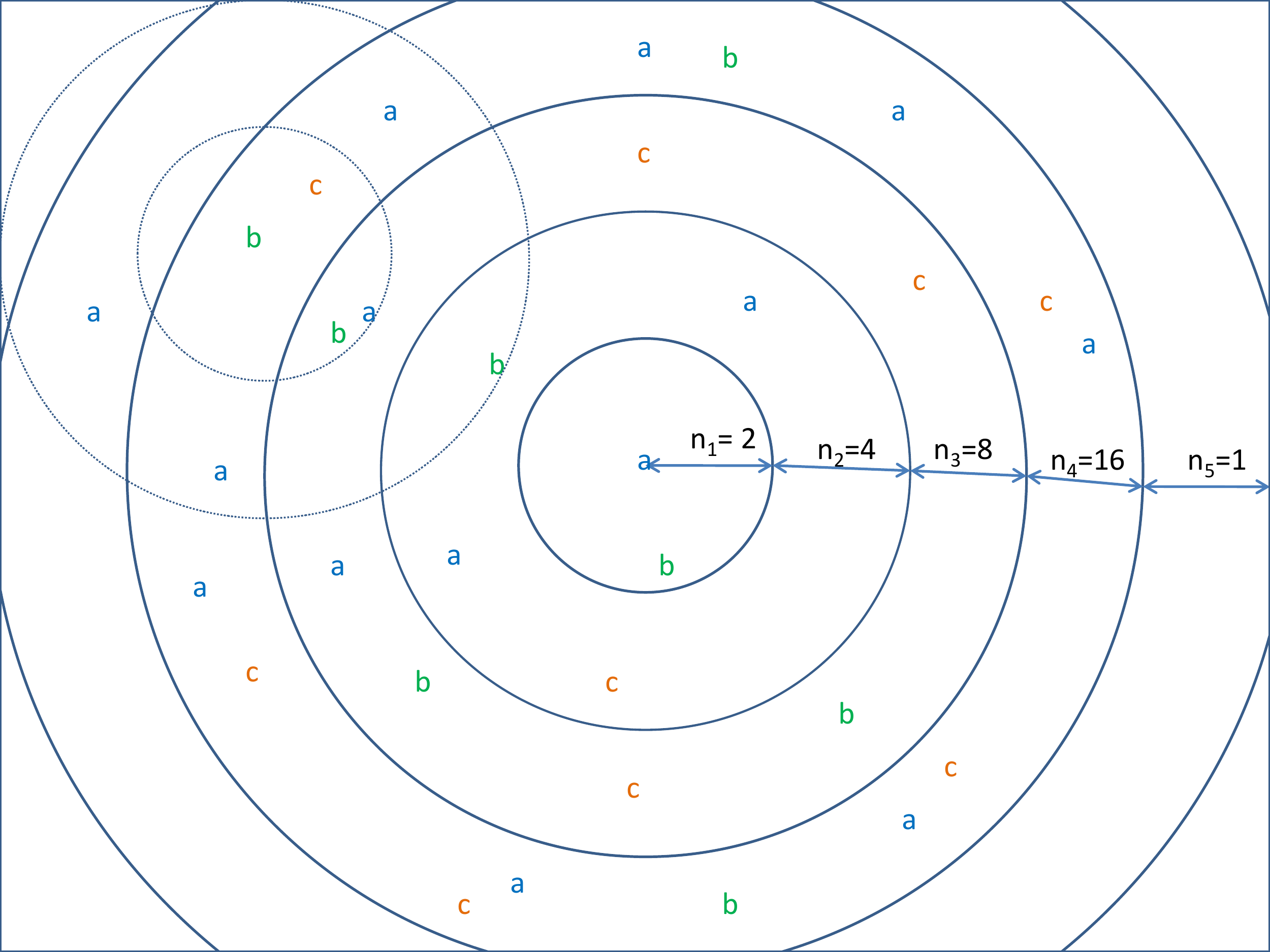}
  \caption{An image with its keypoints feature vectors represented as visual words. Concentric circles are drawn around keypoint $a$ and keypoint $b$. The profile of a keypoint is defined by a set of histograms. Each histogram summarizes the keypoints in a ring. The number of keypoints in a ring doubles as we move outwards from the center.}
  \vspace*{-1mm}
  \label{fig_conc}
  \vspace*{-1mm}
\end{figure*}
\section{Related Work}
\label{sec:related_work}
Svetlana et al.~\cite{Svetlana_Pyramid} proposed to construct a single feature vector for the whole image by concatenating histograms of local features of sub-regions obtained by repeatedly splitting the image at various scales in a principled way (Spatial Pyramid). They used it for full-image categorization. This feature is a global feature vector and is not directly designed for sub-image search. Feature vector may fail to find those matching sub-regions of database images for a given query which get split into multiple regions during feature vector construction. The similarity score of a sub-region of an image with its full-image using this feature vector may be small, depending on the similarity measure used, making it difficult for the true match to be distinguished from false matches. Further, this feature is neither rotation nor translation invariant because of the use of orthogonal split lines. Our profile is computed for each keypoint of an image and matches are retrieved based on keypoint profile similarity rather than full-image similarity. Therefore, our profile is naturally suited to sub-image search. We discuss scale, rotation, and translation invariance of our profile in Section~\ref{sec:prof_creation}.

Weijun et al.~\cite{wang_2008} proposed to spatially cluster local descriptors per image with a bound on the number of points in the cluster and its radius. They represented each cluster as a bag of visual words and each image as a collection of these clusters. Images were ranked based on the count of the clusters that were top-k nearest neighbors of the clusters of a given query image. They achieved a precision of 65\%.
Philbin et al.~\cite{Philbin_2007} clustered the local descriptors to build a visual codebook of size $1$ million and represented each image as a bag of visual words. They used the $L_2$ measure to find candidates and then validated those using LO-RANSACK for restricted sets of transformations. They reported mean average precision of 66\% on specialized dataset of landmark images using 1 million visual words and geometric verification. Yan et al.~\cite{Sukthanka_2004} proposed to generate candidates by nearest neighbor search using Locality Sensitive Hashing~\cite{LSH} and validated using RANSACK for sub-image retrieval for near duplicate image detection and copyright protection. They experimented only on a synthetic dataset. Lowe et al.~\cite{LOWE_SIFT}  performed nearest neighbor search using BBF algorithm~\cite{Beis97shapeindexing} and geometric validation by Hough transform to recognize objects in images.
\section{Profile Creation}
\label{sec:prof_creation}

In this section, we design our new feature vector called \textit{profile} which is created for each keypoint in an image. A profile of a keypoint is a structural representation of the spatial layout of all other keypoints around it. We assume that an image $I$ has been preprocessed and transformed into an orderless bag of visual words. We also know the coordinate of each keypoint detected in an image.  To form the profile, we draw concentric circles around each keypoint $p$ of an image. Figure~\ref{fig_conc} shows concentric circles for keypoint $a$ and keypoint $b$. Each ring is represented as a histogram $h$ of visual words of keypoints lying in it. Profile $\mathbf{H} = (h_1, h_2, \cdots, h_m)$ of a keypoint $p$, where $m$ is the number of rings, is a concatenated list of ring histograms ordered from the center towards the outer rings. The dimension of a profile is directly proportional to the codebook size and the number of rings around a keypoint. The number of points in a ring (called size) increases as we move away from the center, and is defined by  $Size(h_i)=2*Size(h_{i-1})$. The number of points in the first ring is $n_0$ which is user defined and is same for every keypoint's profile across all the images. The $i$th ring contains $2^{i-1}n_0$ points except in the case of the last ring where the required number of points may not exist. It's worth noting that we do not fix the radii of the rings but fix the number of points in a ring, which is a function of $n_0$. As a result, number of rings in the profiles of all the keypoints in a given image will be the same but the radii of the rings may vary. Radii of the rings are based on the spatial density of other keypoints around a given keypoint. An image having more keypoints will have more rings than an image with less keypoints  irrespective of the size of the images. Rings of an image with a high spatial density of keypoints will have narrower rings compared to an image having a low spatial density of keypoints. The profile of each keypoint of an image will differ depending on its position in the image because keypoints captured in the rings of different profiles will be different.\\

\textbf{Similarity between profiles:} The similarity between two profiles $H_i$ and $H_j$,  where the number of rings in $H_i$ is $m$ and in $H_j$ is $n$ respectively, is given by
\begin{eqnarray*}
  Sim(H_i,H_j) &=& \sum_{k=0}^{min(m,n)} e^{-\lambda k} * S_k \\
  S_k &=& Sim(h_{k}(H_i),h_{k}(H_j))
\end{eqnarray*}
Here, $\lambda$ is a decaying parameter learned empirically and is always positive. The similarity between corresponding ring histograms ($S_k$) can be computed using Jaccard's Coefficient or Cosine measure. The distance between two profile is computed as a complement of their similarity. For the purpose of explanation, assume Jaccard's Coefficient as the measure of similarity. For this measure, the maximum value of similarity between two corresponding histograms is $1$. Therefore, the best value of similarity between two profiles is
\begin{eqnarray*}
  Sim_{max}(H_i,H_j) &=& \sum_{k=0}^{min(m,n)} e^{-\lambda k} * 1
\end{eqnarray*}
The distance between two profiles is given by
\begin{eqnarray*}
  D(H_i,H_j) &=& Sim_{max}(H_i,H_j) - Sim(H_i,H_j) \\
             &=& \sum_{k=0}^{min(m,n)} e^{-\lambda k}*(1-S_k)\\
\end{eqnarray*}
This can be represented as a recurrence
\begin{eqnarray*}
    D(H_i,H_j) &=& D(H_i,H_j)_l + \sum_{k=(l+1)}^{min(m,n)} e^{-\lambda k}*(1-S_k)
    \label{distance}
\end{eqnarray*}
where $D(H_i,H_j)_l$ is the distance computed for the first $l$ corresponding ring histograms.
Other similarity measures and their complements can be used to find similarity and distance between two profiles respectively. We found Jaccard's Coefficient as a measure of similarity between histograms to perform better than other methods empirically.

The similarity in the proximity of a keypoint should be weighed higher but should be weighed less as we move away to decimate the effect of noise and overfitting in matching. This is the reason for the exponentially decaying aggregation of similarity between corresponding rings of profiles and keeping increasingly more keypoints in the rings away from the center.
\begin{figure}[t]
  \centering
      \includegraphics[height=1.5in,width=0.45\textwidth]{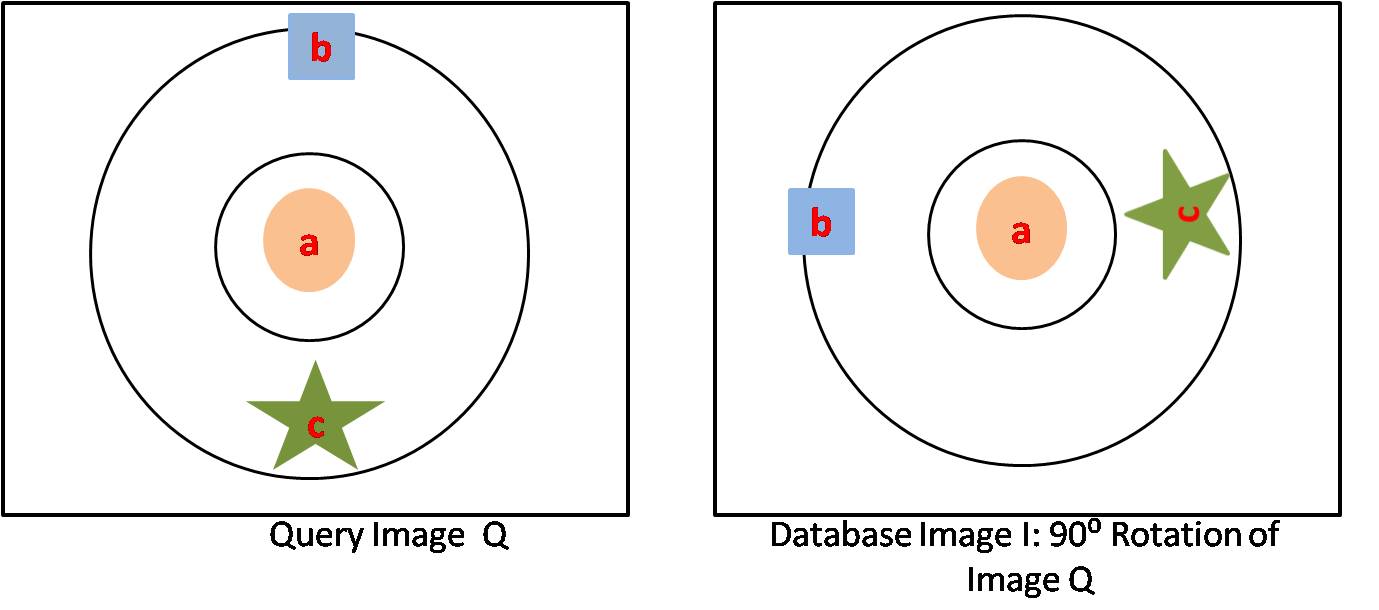}
  \caption{The profile of keypoint $a$ in query image Q is same as its corresponding profile in rotated image I.}
  \vspace*{-1mm}
  \label{fig:rot_invariance}
  \vspace*{-3mm}
\end{figure}

A keypoint is localized in an image and its local descriptor captures the property of its small neighborhood in the image. Locality of a keypoint makes it naturally fit for sub-image search. Our concentric profile captures spatial layout of keypoints of the whole image  with respect to a given keypoint. The profile of a keypoint gives it the structured global view of the whole image making it semantically richer than the keypoint itself. Therefore, the profile of a keypoint can distinguish between a true match and a false match effectively. A sub-image search based on profiles would have higher accuracy than bag of visual words based search.

\begin{figure}[t]
  \centering
      \includegraphics[width=0.45\textwidth]{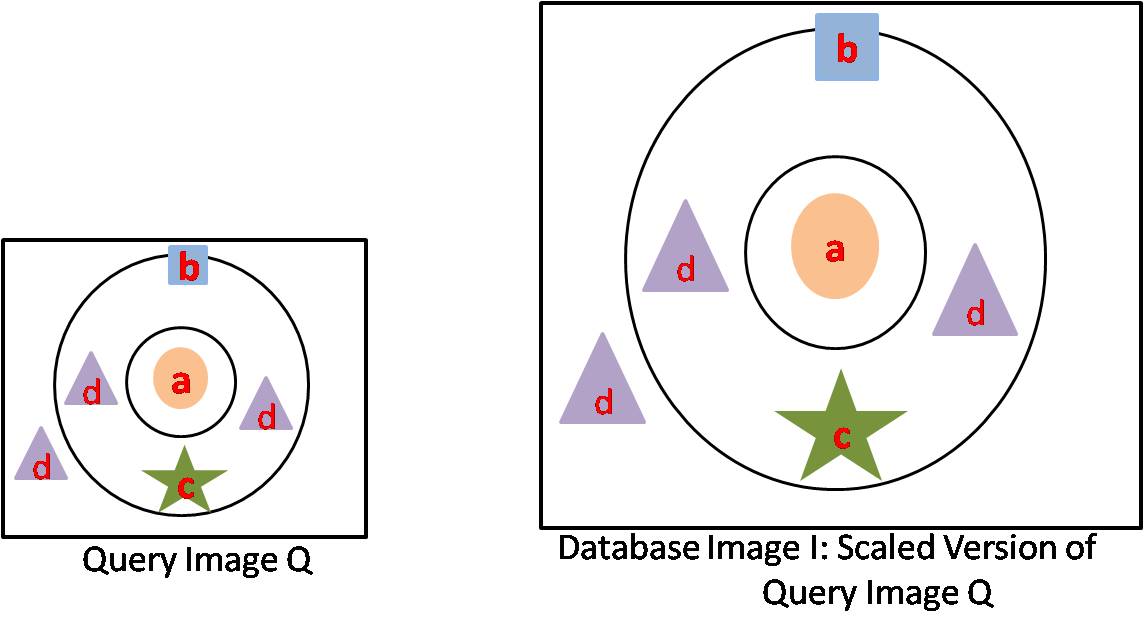}
  \caption{The profile of keypoint $a$ in image Q and image I remain the same in spite of scaling. The ring radius has increased in image I with an increase in scale.}
  \vspace*{-1mm}
  \label{fig:scale_invariance}
  \vspace*{0mm}
\end{figure}

\textbf{Profile Vector Robustness:} Visual words which form our profile's histogram are obtained from local descriptors. It should be noted that although local descriptors like SIFT provide invariance to affine changes for similarity search, feature vectors created by spatial division of image around a keypoint may compromise invariance to affine changes. Our profile vector retains invariance to rotation, scaling, translation, and occlusion. We perform image division only in a radial direction. Therefore, our method yields same concentric profile for a given keypoint irrespective of the rotation of an image making it rotation invariant. We see in Figure~\ref{fig:rot_invariance} that the profile of keypoint $a$ in query image $Q$ remains same in rotated image $I$. The scaling of an image alters the relative distance between points, as seen in image $I$ of Figure~\ref{fig:scale_invariance}. Therefore, a fixed radius concentric division will fail to provide scale invariance even though the local descriptors are scale invariant. We keep equal number of keypoints in rings while constructing the profile of a given keypoint irrespective of the size and scale of an image. This technique generates the same profile for a given keypoint in two images irrespective of the scale. The profile of keypoint $a$ of image $Q$ in Figure~\ref{fig:scale_invariance} is the same as the profile of keypoint $a$ in the scaled image $I$. Therefore, our profile feature remains invariant to scaling. Our search methodology automatically preserves translation invariance and provides robustness to occlusion as discussed in Section~\ref{sec:algorithm}.

\section{Sub-Image Search}
\label{sec:algorithm}
\begin{figure}[t]
  \centering
      \includegraphics[width=0.45\textwidth]{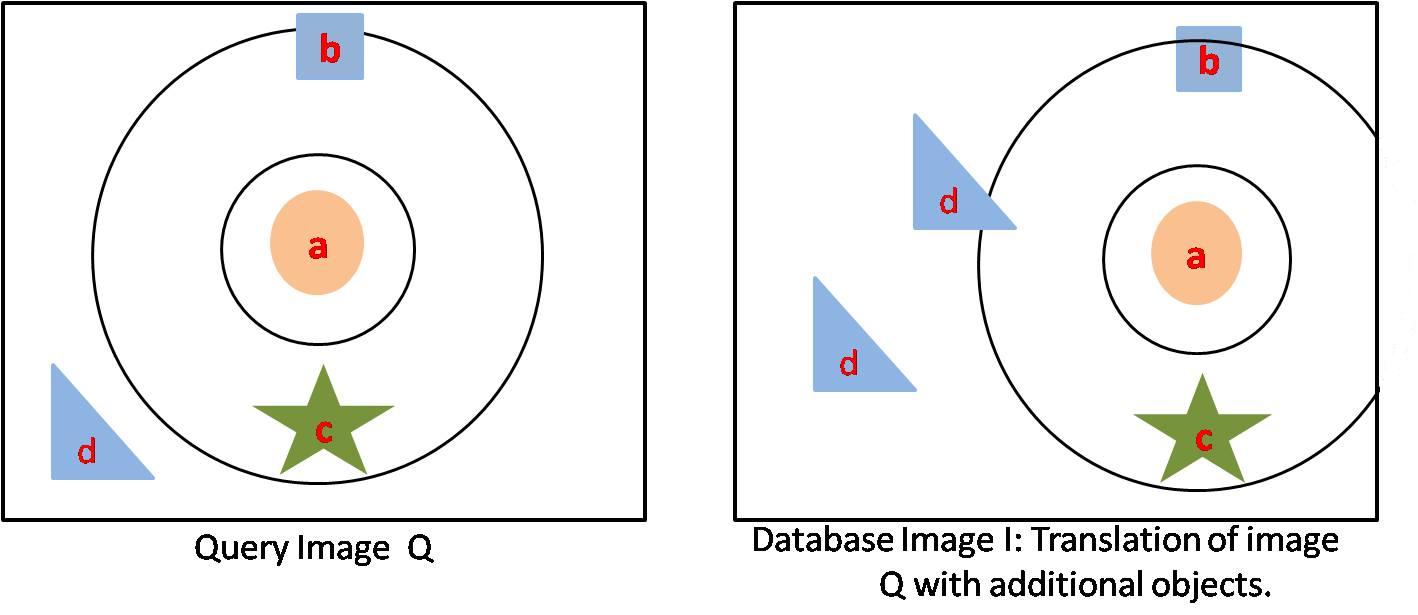}
  \caption{Keypoint $a$ is translated in image I. The profile of keypoint $a$ in image I has high similarity to its profile in image Q.}
  \vspace*{-1mm}
  \label{fig:Trans_invariance}
  \vspace*{0mm}
\end{figure}

\begin{figure}[t]
  \centering
      \includegraphics[width=0.45\textwidth]{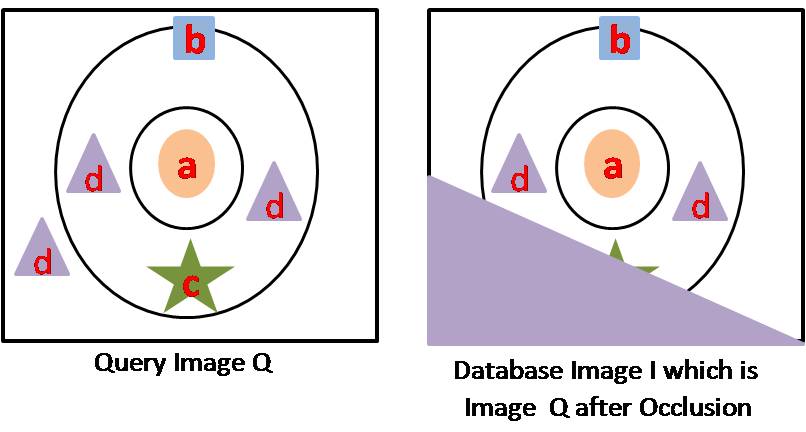}
  \caption{The profile of keypoint $a$ in image I has high similarity with profile of $a$ in image Q even after occlusion.}
  \vspace*{-1mm}
  \label{fig:Occ_invariance}
  \vspace*{0mm}
\end{figure}

In this section, we discuss the method used to compute the similarity between a pair of profiles. This similarity is used for sub-image search. Similar keypoints will be extracted from a query image $Q$ and a matching sub-region of a database image $I$ by local descriptor methods. Therefore, the keypoints captured in the more weighted histograms of the profile of a keypoint $q$ in a query image $Q$ would be same to the profile of a keypoint $p$ in the matching sub-region of a database image $I$. This will give high similarity score between the profiles. We exploit this property to retrieve best matching sub-regions from the database. All the images, represented as bags of words, are converted into bags of profiles. Query $Q$ is also processed to generate a bag of profiles. We search the best matching profile $p_i$ in a database for each query profile $q_j$ and finally, choose the highest scoring pair ($q_j$, $p_i$) among these. The sub-region around the keypoint $p_i$ of the highest scoring pair is the best matching image. If $Q$ has $m$ profiles and $N$ is the total number of profiles in the dataset then the best matching sub-image is the region around $i$th profile where $i$ is obtained by
\begin{eqnarray*}
   \arg_{i}\max_{j \le m}\left\{\max_{i \le N}Sim(H_{q_j},H_{q_i})\right\}
\end{eqnarray*}
 Our search algorithm inherently provides translation invariance because it searches for the best matching profile. We create a profile for each keypoint of an image. Since, local descriptors are translation invariant, relatively similar keypoints will be detected in a matching sub-region of a translated image and the given query. Therefore, profiles of the query will have high similarity with the profiles of the keypoints detected in the matching sub-region compared to the profiles of random keypoints and our algorithm will successfully find these translated matches. Image $I$ in Figure~\ref{fig:Trans_invariance} has translation of few objects in image $Q$ as well as additional new objects. We see that the concentric profile of keypoint $a$ in image $I$ is very similar to the corresponding profile of $a$ in image $Q$. Our algorithm will also find a partially occluded but true matching sub-region of a database. Profile of a keypoint detected in the preserved part of the matching sub-region will have relatively high similarity score with some keypoint profile of the given query. Profile of keypoint $a$ in image $I$ of Figure~\ref{fig:Occ_invariance} will have high similarity with the profile of keypoint $a$ in image $Q$. Therefore, the true matching sub-region will rank higher on similarity score compared to random profiles.

\section{Experimental Evaluation}
\label{sec:experiment}
In this section, we present a comparative study with a state of art technique using the author's dataset. Next, we describe our own dataset and perform experiments to show that our profile based approach yields high precision for sub-image search with a small visual codebook.
\subsection{Comparative Evaluation}
\begin{figure}[t]
    \vspace*{-2mm}
	\begin{center}
		\includegraphics[width=0.8\columnwidth]{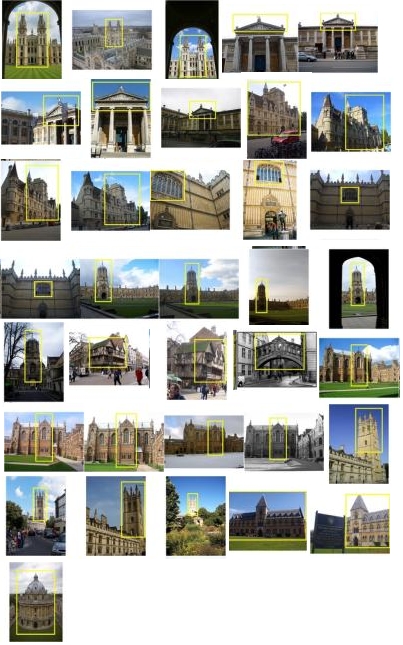} 
		\vspace*{-1mm}
		\caption{36 sub-image queries marked with yellow box from Oxford Dataset.}
		\label{fig:queries}
	\end{center}
	\vspace*{-8mm}
\end{figure}

We applied our profile feature based search on the Oxford dataset provided by Philbin et al.\footnote{http://www.robots.ox.ac.uk/~vgg/data/oxbuildings/index.html}~\cite{Philbin_2007}. This dataset has bag of visual words representation of 5,062 high resolution (1024 $\times$ 768) images of 11 different Oxford buildings collected from Flickr. It has 55 queries with ground truth for comparative validation. Philbin et al.~\cite{Philbin_2007} used the $L_2$ distance on bag of words model to rank all the images for each query and then computed the average precision. They reported mean average precision of 64.5\% without geometric validation and mean average precision of 66\% with geometric validation. We chose the top-36 most challenging queries which were better representative of sub-image search as seen in Figure~\ref{fig:queries}, and specifically left out the queries which were more of a full image search. For a given $k$, we retrieved top-k similar images for each of the 36 queries from this dataset. We computed the ratio of true results retrieved to the total images retrieved (precision). We averaged the precision over all the $36$ queries. We also retrieved top-$k$ results for each query by just computing the $L_2$ distance on bag of words model as proposed by Philbin et al. and computed mean precision. We present the comparative percentage precision in Table~\ref{tab:pisr_oxf_prec} for varying $k$ from 1 to 10. For a profile based search, we can see that the top-$5$ results have mean precision as high as \textbf{94}\% and it drops to only \textbf{86}\% for top-$10$ results. For a bag of words based search, mean precision drops to 77\% for top-$10$ results. A typical user generally looks at top few results of a query returned by a search engine. We see that our profile based feature vector returns highly accurate results for challenging queries on the landmark images in top-$10$ compared to the state of the art method.

\begin{table}[t!]
	\begin{center}
		\begin{tabular}{|c|p{2.5cm}|p{2.5cm}|}
                  \hline
                   Rank k & \% precision for profile based search  & \% precision for bag of words using $L_2$ based search  \\
                  \hline
                   1 & 100  & 100\\
                  \hline
                   2 & 100 & 98\\
                  \hline
                   3 & 96  & 93 \\
                  \hline
                   4 & 95 & 92\\
                  \hline
                   5 & 94  & 90\\
                   \hline
                    6 & 94 & 89\\
                  \hline
                   7 & 92  & 86\\
                   \hline
                   8 & 90  & 83\\
                  \hline
                   9 & 87  & 80\\
                   \hline
                   10 & 86 & 77\\
                  \hline
		\end{tabular}
		\vspace*{0mm}
		\caption{Comparative precision for 36 Oxford building queries for varying $k$.}
		\label{tab:pisr_oxf_prec}
	\end{center}
	\vspace*{-8mm}
\end{table}
\subsection{Dataset Preparation} 
\begin{figure}[t]
    \vspace*{-2mm}
	\begin{center}
		\includegraphics[width=0.95\columnwidth]{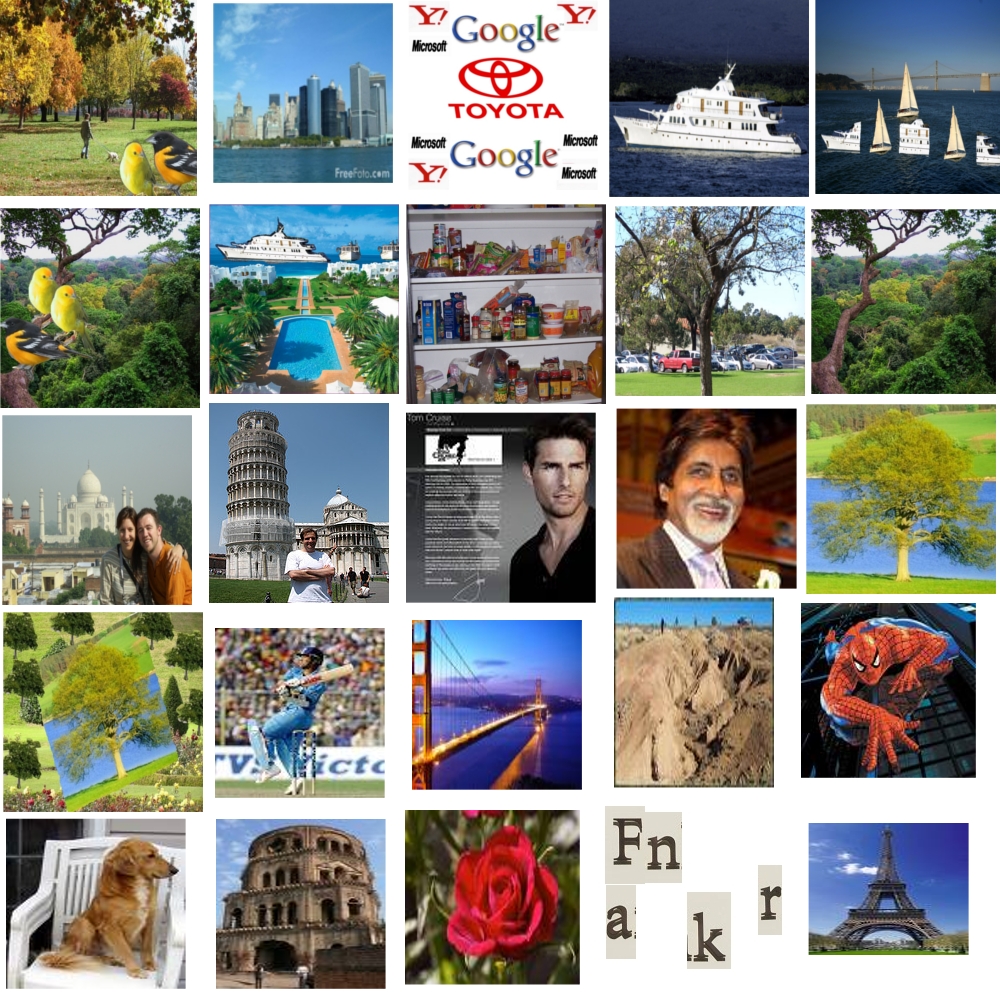}
		\vspace*{-1mm}
		\caption{A snapshot of our experimental dataset which has many variations and is a close representative of a natural image dataset.}
		\label{fig:data_snapshot}
	\end{center}
	\vspace*{-5mm}
\end{figure}

We downloaded natural images from Flickr. We also manually photographed a large number of scenes under varying conditions. We chose $35$ random queries from this real dataset of natural images for our experiments. We synthetically created some test images to put our method to even tougher challenges. We randomly chose $17$ sub-images from the natural images and embedded them into other large natural images. We also added varying noise levels to these images as discussed in~\cite{MENG_TRANSFORM}. Some of the operations to add noise were rotation, scaling, shearing, gaussian blur, and averaging noise. We also split a sub-image and scattered its fragments into other images. Finally, we obtained a dataset of $1000$ images ($800$-natural and $200$-synthetic). A snapshot of the dataset is shown in Figure~\ref{fig:data_snapshot}. We used a total of $52$ queries ($35$-natural and $17$-synthetic).

Next, we created bag of words representation of each image. We extracted covariant regions~\cite{comp_affine} from each image and the corresponding $128$ dimensional SIFT vector. We clustered the feature vectors by picking $500$ random centers using k-means~\cite{k-means} clustering. We chose k-means clustering because it is not only simple but embarrassingly parallelizable. A naive computation of k-means has complexity of O($mnt$) where $m$ is the number of cluster centers, $n$ is number of data points, and $t$ is the number of iteration for k-means convergence. This can be reduced by using metric property~\cite{Fast_Kmeans}. We repeated this process $10$ times to choose the cluster set which has the minimum average sum of the squared error or scatter. We assigned the symbol to each cluster. We mapped each SIFT feature vector to the cluster symbol to which it belonged. This gave us a bag of words representation of an image. For creating profiles, we chose $n_0$=50 points in the first ring and used $\lambda$=1/3 as decaying parameter for aggregation. We found the Jaccard's Coefficient to yield better results than other distance measures and used it to compute similarities between ring histograms of our profiles.
\subsection{Precision Test} 
\begin{figure}[t]
  \centering
      \includegraphics[width=0.45\textwidth]{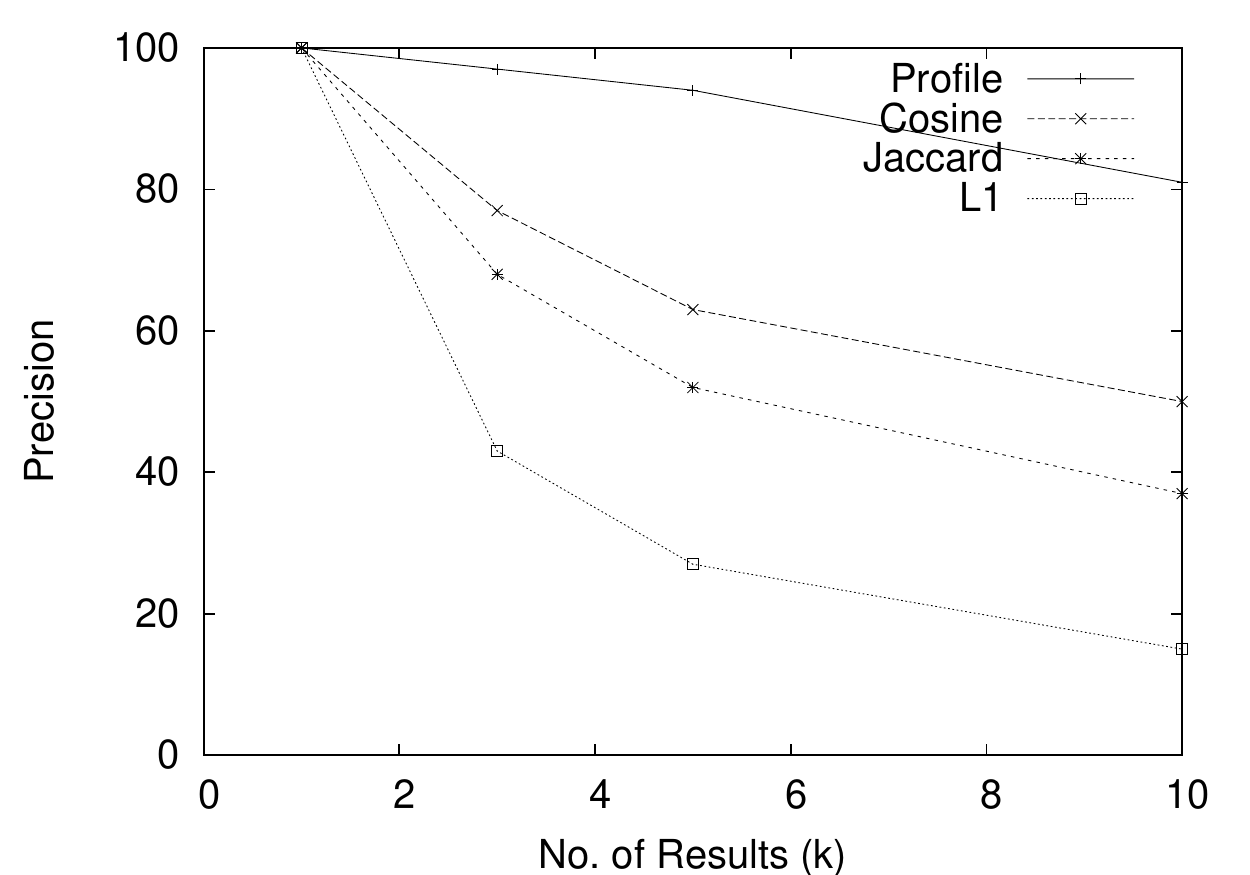}
      \vspace*{-1mm}
  \caption{Comparison of conventional methods, using different distance measures without tf-idf weighting, with profile based approach.}
  \label{fig_non_weighted}
  \vspace*{-3mm}
\end{figure}

\begin{figure}[t]
  \centering
      \includegraphics[width=0.45\textwidth]{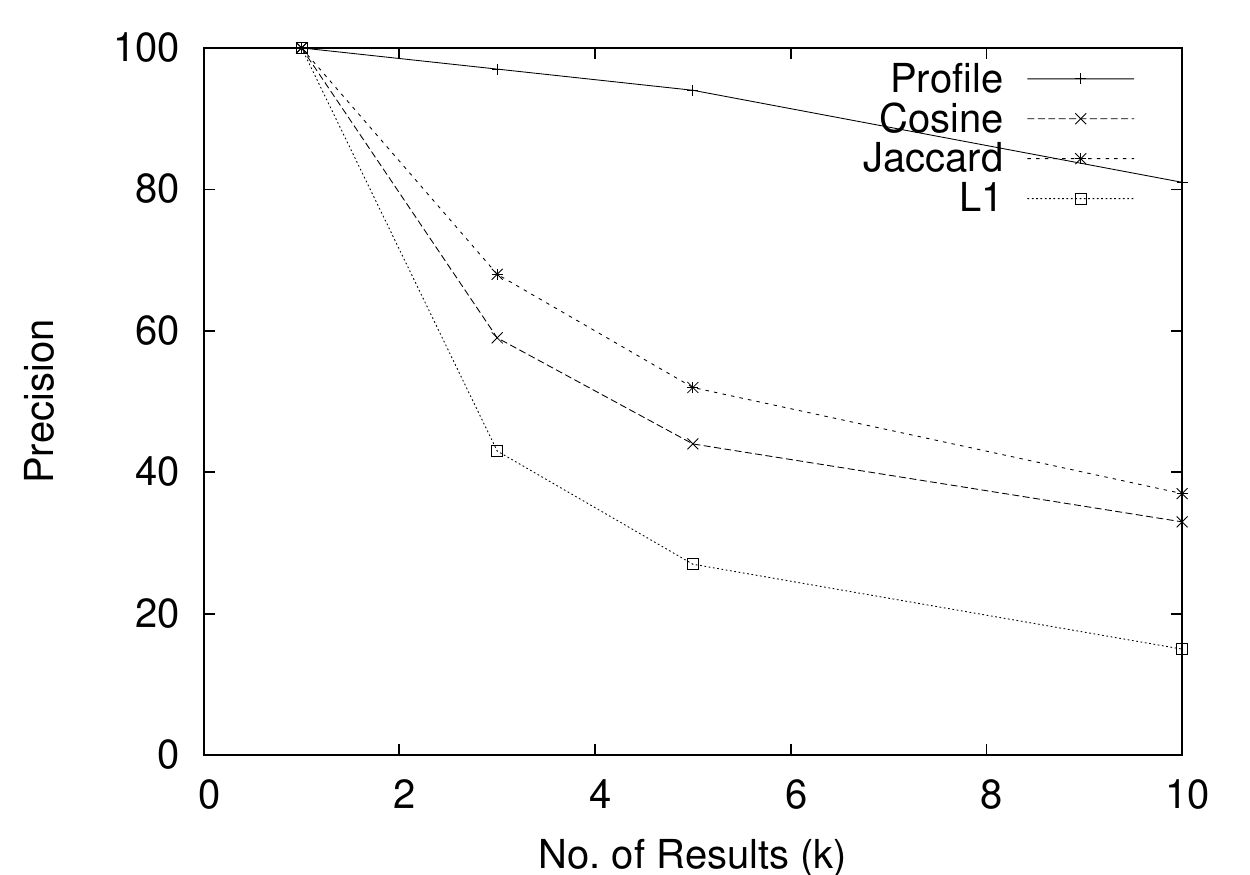}
  \vspace*{-1mm}
  \caption{Comparison of conventional methods, using different distance measures and tf-idf weighting, with profile based approach.}
  \label{fig_weighted}
  \vspace*{-3mm}
\end{figure}

We computed precision of top-k results obtained using our profile based search and compared it against the precision of conventional methods which compute similarity only on the bag of words without taking spatial relationships into account. We used Cosine, $L_1$, and Jaccard's Coefficient as the measure of similarity for conventional methods. We considered both the schemes of standard tf-idf weighting~\cite{tf-idf} and without tf-idf weighting of visual words for the candidate generation approach. In a tf-idf approach, commonly occurring visual words are weighed less as they are less discriminative. We did not consider the tf-idf weighting of visual words for our profile based approach. Figure~\ref{fig_non_weighted} and Figure~\ref{fig_weighted} show the comparative precision of various methods for the non-weighted and the weighted cases, respectively. We find that our approach yields 81\% precision rate for top-10 results. Without tf-idf weighting, conventional method with cosine measure gives the best precision of 50\% which is $31$\% less than profile based method. With tf-idf weighting, conventional method with Jaccard's coefficient gives the best precision of 39\% for top-10 results. Search with tf-idf weighting achieves less precision compared to the search without tf-idf weighting for conventional methods. We also experimented by varying $\lambda$ for our profile similarity and achieved more than 20\% higher precision for every $ \lambda \leq 1$ compared to conventional methods. We also experimented by weighting the symbols with the area of covariant regions but achieved less precision. We see that our method achieves higher precision than conventional methods used for candidate generation on a small codebook size and without any geometric verification.

\subsection{Visual Results}
\begin{figure*}[t]
	\begin{center}
		\begin{tabular}{|c|c|c|c|c|c|c|}
              \hline
              S. No. & Query & 1st & 2nd & 3rd & 4th & 5th\\
              \hline
              1 &
              \includegraphics[height=\figheight,width=\figwidth]{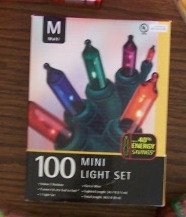} &
              \includegraphics[height=\figheight,width=\figwidth]{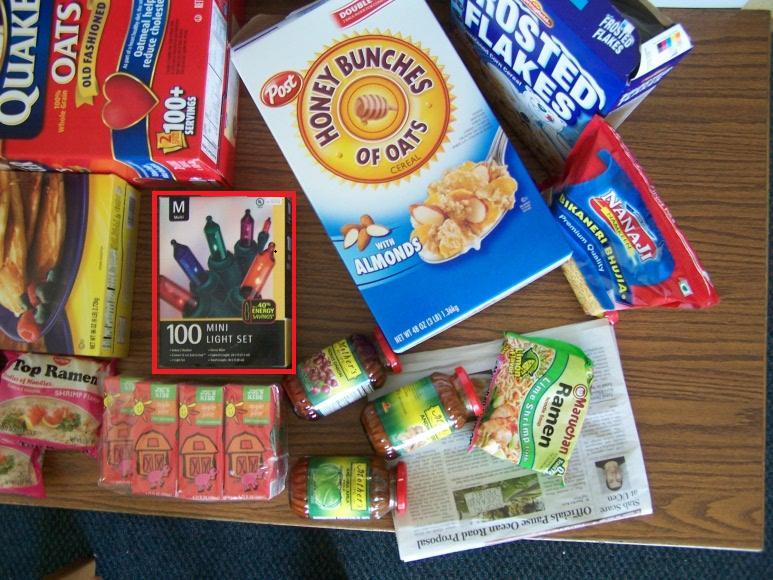}&
              \includegraphics[height=\figheight,width=\figwidth]{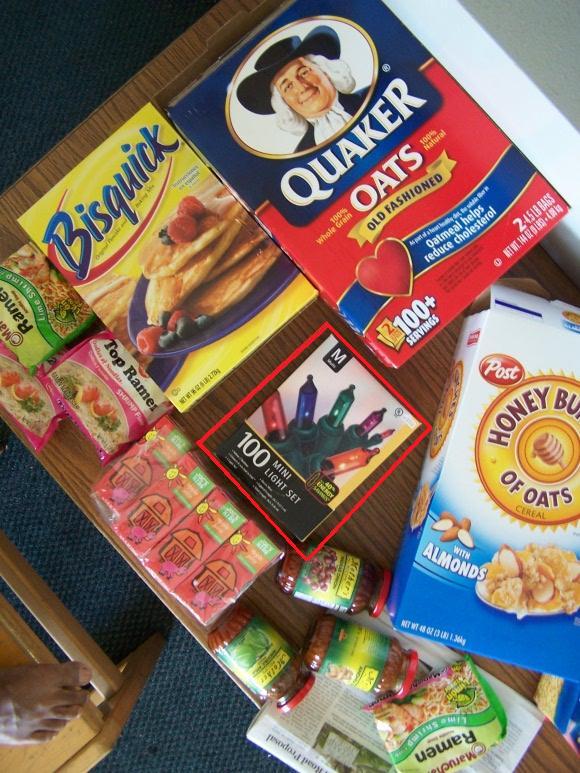} &
              \includegraphics[height=\figheight,width=\figwidth]{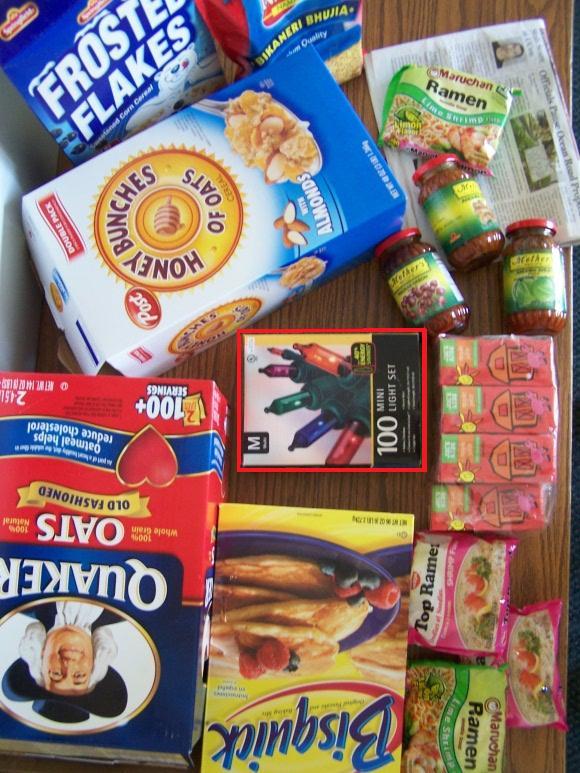} &
              \includegraphics[height=\figheight,width=\figwidth]{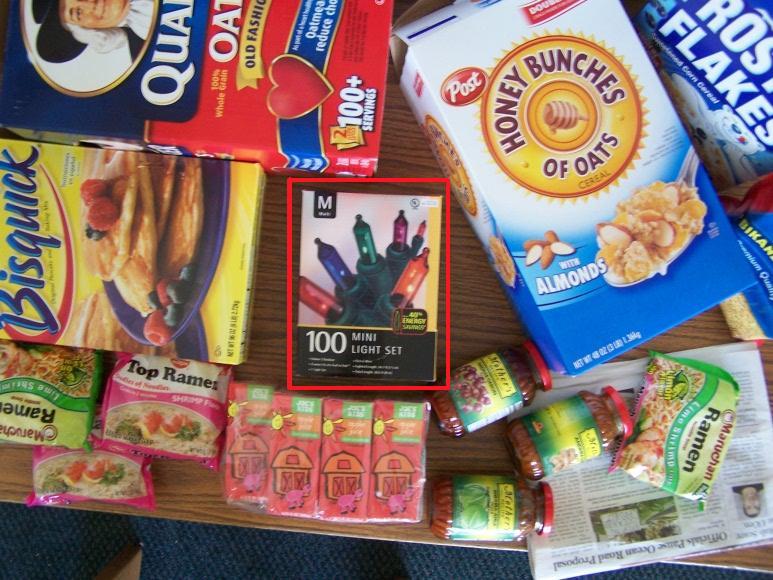}&
              \includegraphics[height=\figheight,width=\figwidth]{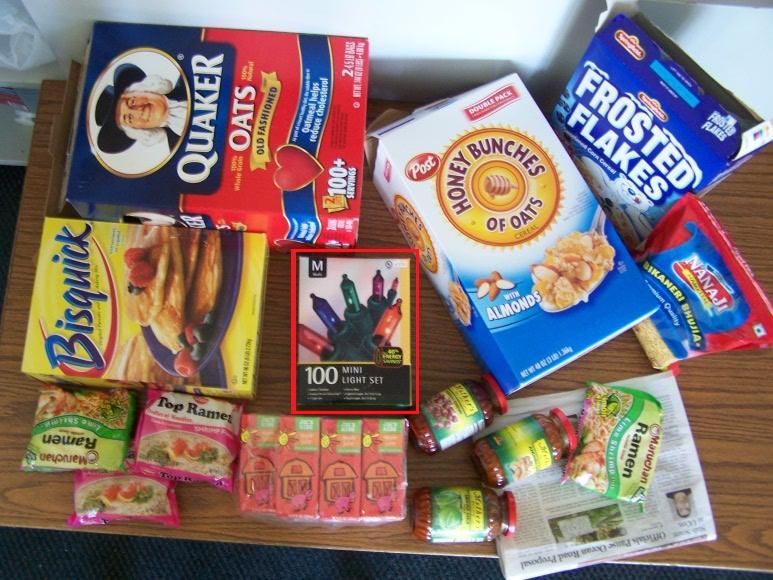}\\
              \hline
              2 &
              \includegraphics[height=\figheight,width=\figwidth]{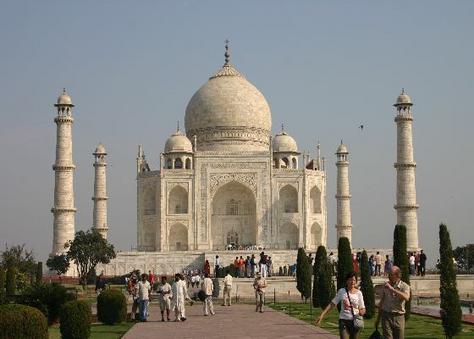} &
              \includegraphics[height=\figheight,width=\figwidth]{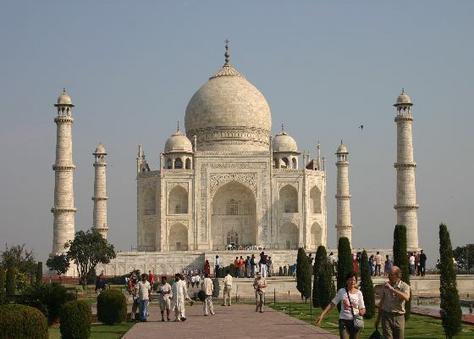}&
              \includegraphics[height=\figheight,width=\figwidth]{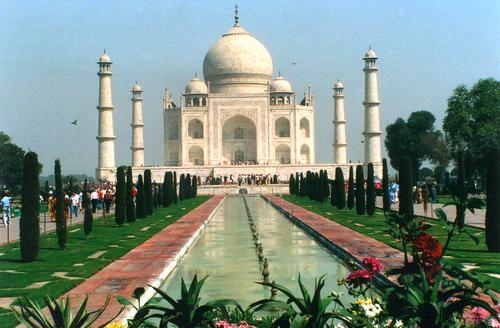} &
              \includegraphics[height=\figheight,width=\figwidth]{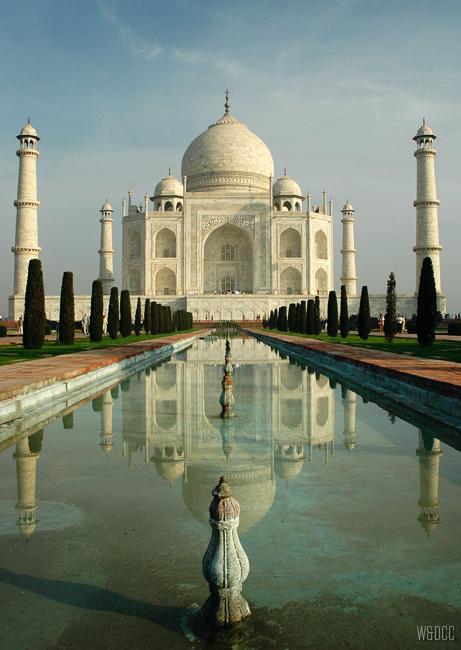} &
              \includegraphics[height=\figheight,width=\figwidth]{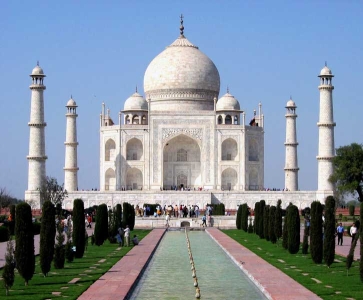}&
              \includegraphics[height=\figheight,width=\figwidth]{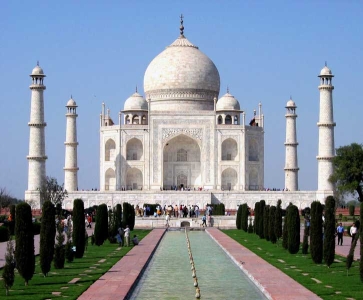}\\
              \hline
              3 &
              \includegraphics[height=\figheight,width=\figwidth]{357} &
              \includegraphics[height=\figheight,width=\figwidth]{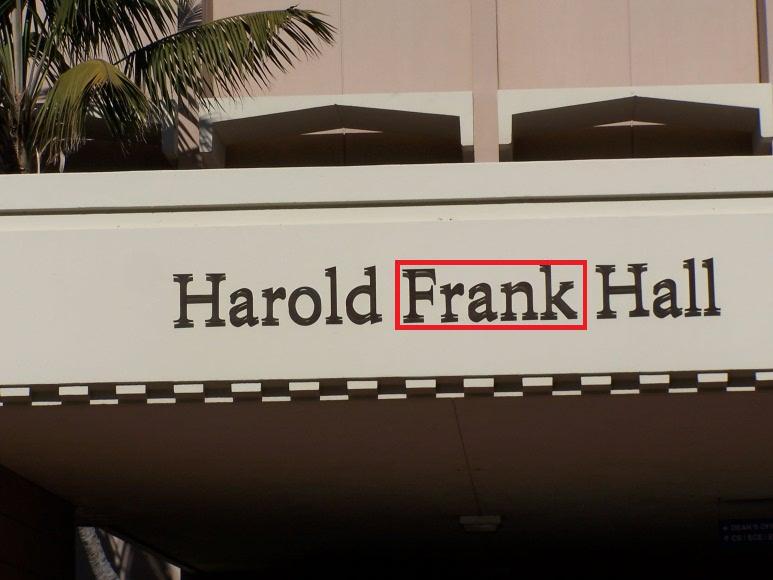}&
              \includegraphics[height=\figheight,width=\figwidth]{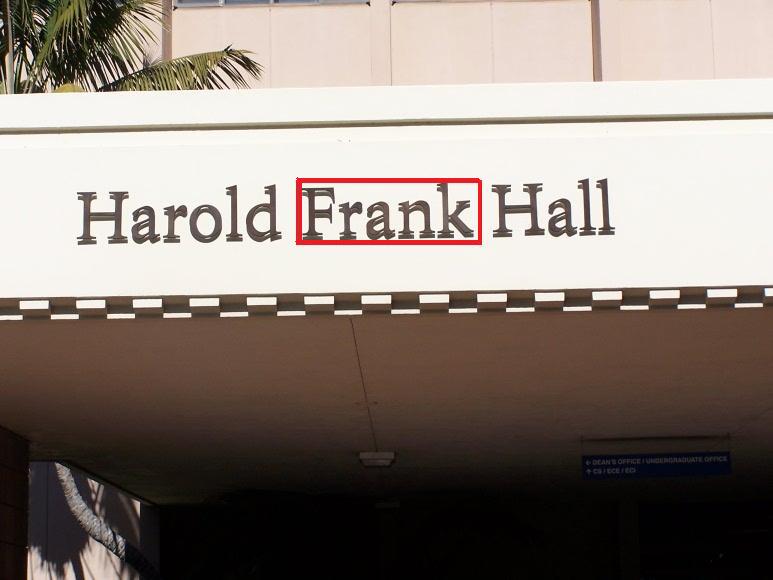} &
              \includegraphics[height=\figheight,width=\figwidth]{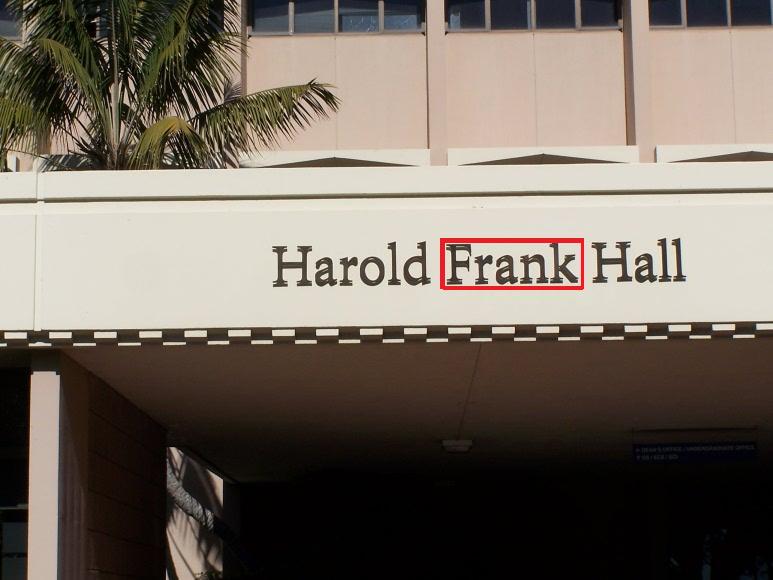} &
              \includegraphics[height=\figheight,width=\figwidth]{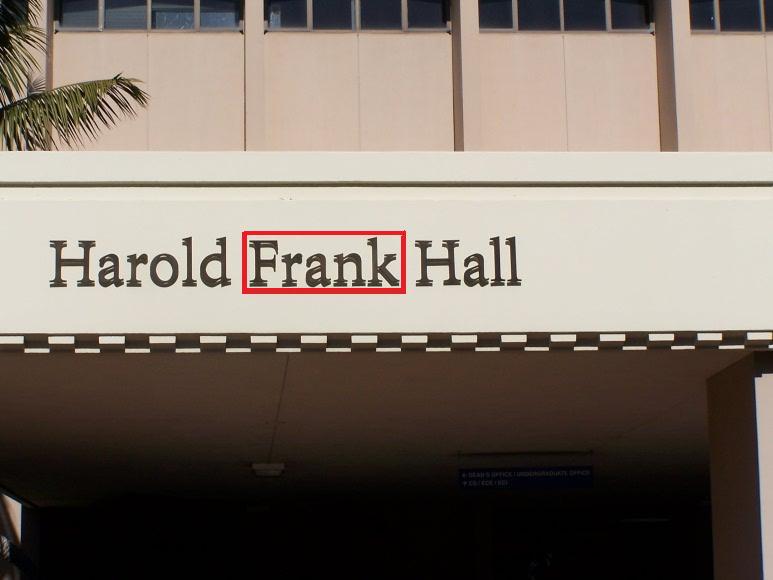}&
              \includegraphics[height=\figheight,width=\figwidth]{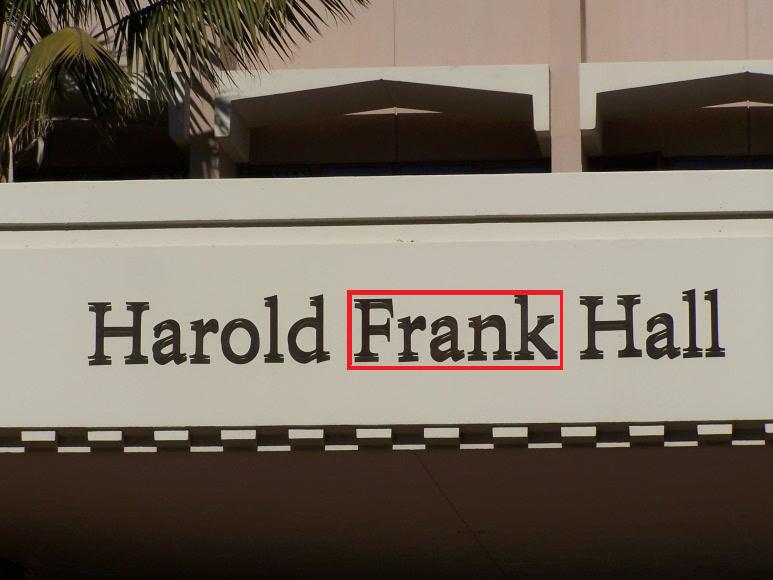}\\
              \hline
              4 &
              \includegraphics[height=\figheight,width=\figwidth]{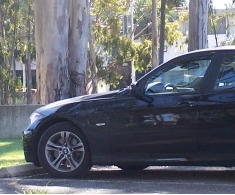} &
              \includegraphics[height=\figheight,width=\figwidth]{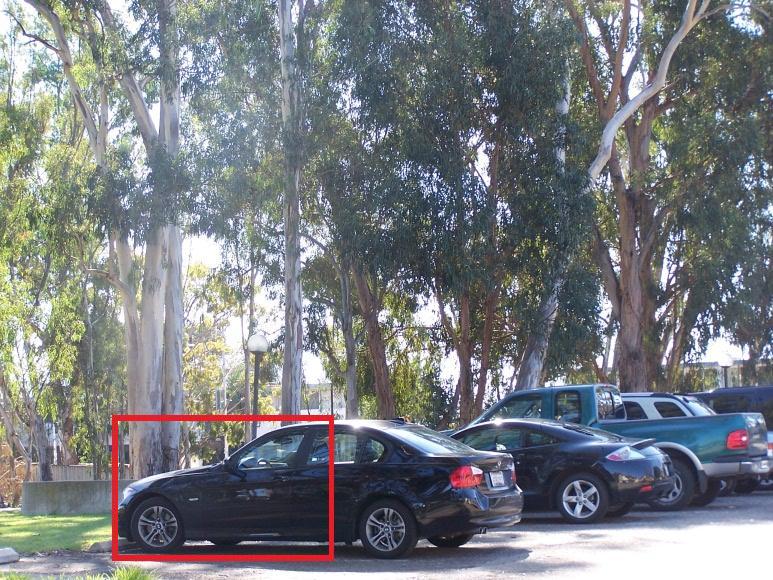}&
              \includegraphics[height=\figheight,width=\figwidth]{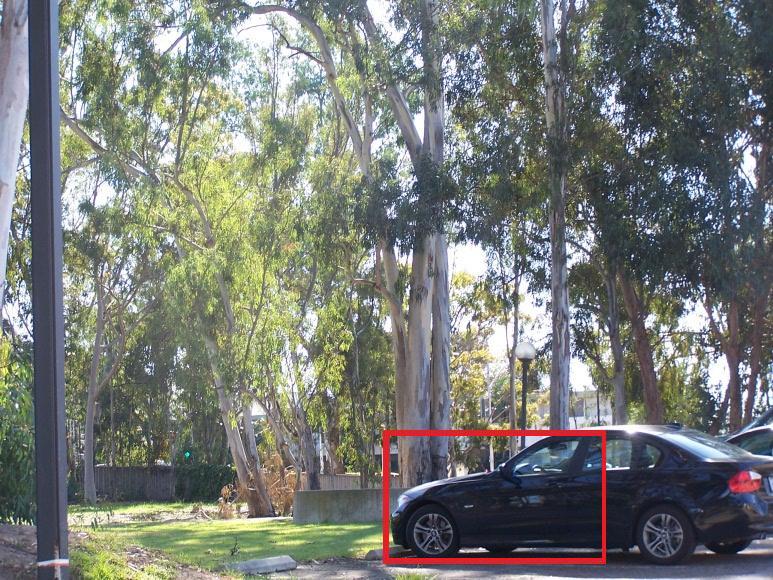} &
              \includegraphics[height=\figheight,width=\figwidth]{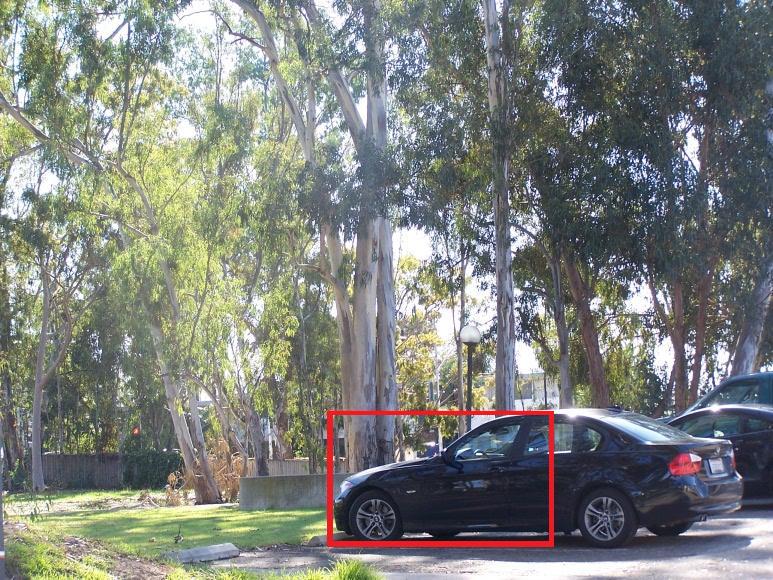} &
              \includegraphics[height=\figheight,width=\figwidth]{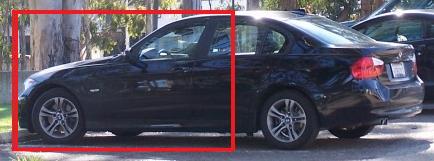}&
              \includegraphics[height=\figheight,width=\figwidth]{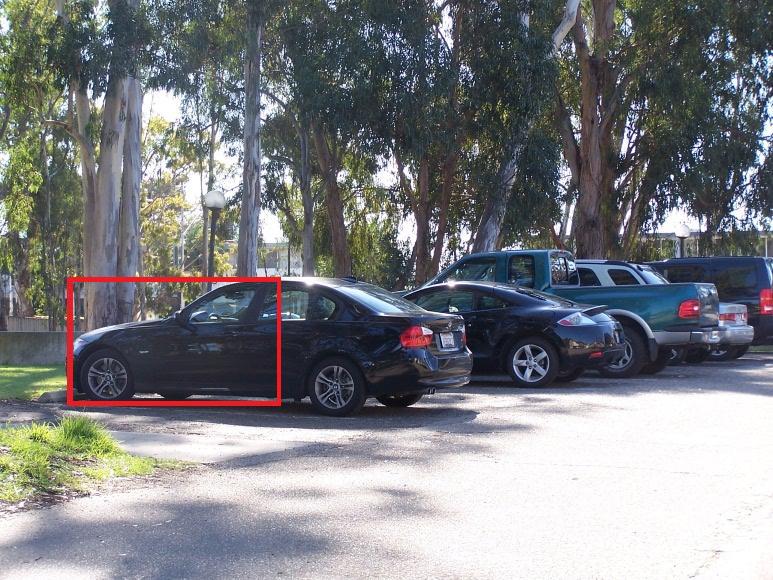}\\
              \hline
         \end{tabular}
  \caption{Top-5 results for 4 real queries over general image dataset using profile based search.}
  \label{tab_result}
  \vspace*{-6mm}
  \end{center}
\end{figure*}

We present the top-5 visual results for $4$ real queries from our search in Figure~\ref{tab_result}. Our profile based approach retrieves high quality results irrespective of the kind of noise present in the dataset. We outline the matching sub-region in a result image with red box.  We got all true matches in the top-5 results for the $3$rd query which got pathological result using conventional methods shown in Figure~\ref{fig:path_1}. We also verified for the 2nd pathological case shown in Figure~\ref{fig:path_2} and found that the result was returned in top-10 results and all the better scoring images were true matches.
\section{Conclusion}
In this paper, we developed a simple but effective profile based feature vector that captures the spatial relationships between keypoints of an image. Inclusion of spatial layout in feature vector improves the search result dramatically without the need of geometric verification.  We proposed a method to measure the similarity between profiles. Our technique produced higher accuracy on landmark images compared to the state of the art method.  We evaluated our technique on a mixture of synthetic and real natural images dataset over 52 queries and obtained a 81\% precision for top-$10$ matches using a small code book size of $500$. This was 31\% higher than the conventional methods.

{\footnotesize
\bibliographystyle{abbrv}
\bibliography{PISR,GeoClusteringNoGeoTag}
}
\end{document}